# GlobalMind: Global Multi-head Interactive Self-attention Network for Hyperspectral Change Detection

Meiqi Hu, *Graduate Student Member, IEEE,* Chen Wu, *Member, IEEE,* and Liangpei Zhang, *Fellow, IEEE*

*Abstract*—High spectral resolution imagery of the Earth's surface enables users to monitor changes over time in fine- grained scale, playing an increasingly important role in agriculture, defense, and emergency response. However, most current algorithms are still confined to describing local features and fail to incorporate a global perspective, which limits their ability to capture interactions between global features, thus usually resulting in incomplete change regions. In this paper, we proposed a Global Multi-head INteractive self-attention change Detection network (GlobalMind) to explore the implicit correlation between different surface objects and variant land cover transformations, acquiring a comprehensive understanding of the data and accurate change detection result. Firstly, a simple but effective Global Axial Segmentation (GAS) strategy is designed to expand the self-attention computation along the row space or column space of hyperspectral images, allowing the global connection with high efficiency. Secondly, with GAS, the global spatial multi-head interactive self-attention (Global-M) module is crafted to mine the abundant spatial-spectral feature involving potential correlations between the ground objects from the entire rich and complex hyperspectral space. Moreover, to acquire the accurate and complete cross-temporal changes, we devise a global temporal interactive multi-head self-attention (GlobalD) module which incorporates the relevance and variation of bi-temporal spatial-spectral features, deriving the integrate potential same kind of changes in the local and global range with the combination of GAS. We perform extensive experiments on five mostly used hyperspectral datasets, and our method outperforms the state-of-the-art algorithms with high accuracy and efficiency.

## I. INTRODUCTION

Remote sensing is a powerful tool in Earth observation and monitoring the evolution of land use [1]–[3], widely applied in urbanization [4], agriculture [5], [6], nature disaster [7], climate change [8]. Change detection (CD) aims at detecting the changes of two remote sensing (RS) images acquired on different time at the same location [9]–[11]. CD has long been a fundamental task in remote sensing, where variant changes origin from whether natural land cover or artificial ground objects alteration may affect the living of human being or provide essential information for decision-makers [12]–[14]. Currently, CD has been successfully used in global ecological monitoring [15], land-cover and land usage [16], [17], and disaster response [18], [19]. Hyperspectral image (HSI) consists of hundreds of spectral bands with high spectral resolution, which turns to be one of the key trends of future satellite imaging. HSI produces remarkably detailed information of the Earth's surface, beneficial for fine-grained classification [20], [21], subpixel unmixing [22], [23], anomaly and target detection [24], [25], and change detection [26], [27]. By leveraging the capability of HSI, hyperspectral change detection (HCD) provides unprecedent potential to discriminate the subtle and detailed changes.

With the framework of deep neural networks, HCD has made significant progress in recent years [28]–[30], where most of existing works are devoted to coping with the challenges raised by the high-dimensional spectral space and extensive spectral heterogeneity of rich and redundant hyperspectral imaging [31]–[33]. In order to obtain more accurate change detection results, many current algorithms rely on convolutional networks and RNN networks to extract multiscale spatial-spectral features and discriminative temporal change information. The representative GETNET [34] proposed a general end-to-end convolutional neural network to accomplish the rich feature representation from designed pixel-wise mixed affinity matrix. In SALA [35], Hou et al. put forward a novel distance function to make full use of spatial information by using multiple morphological profiles, smoothing out the noise in change detection result. Except for local spatial information, detailed spectral information in the super abundant channel dimension is also considered [36], [37]. Especially, SSA-SiamNet has incorporated spatial attention and spectral attention to emphasize discriminative channels and locations and suppress less informative ones [37]. However, fine spatial-spectral information alone cannot satisfy the requirements of precise change detection tasks. The rich and diverse land cover on the Earth's surface demands multi-scale feature representation for more accurate interpretation [38], [39]. To deal with the fixed kernel limitation, DMPs2raN [40] presented a deep multiscale pyramid network enhanced with spatial–spectral residual attention to mine multilevel and multiscale spatial–spectral features. A multilevel encoder–decoder attention network (ML-EDAN [41]) was proposed to extract hierarchical features with the designed contextual-information

This work was supported in part by the National Key Research and Development Program of China under Grant 2022YFB3903300 and 2022YFB3903405, and in part by the National Natural Science Foundation of China under Grant T2122014, 61971317, 62225113 and 42230108. (*Corresponding author: Chen Wu.*)

M. Hu, C. Wu and L. Zhang are with the State Key Laboratory of Information Engineering in Surveying, Mapping and Remote Sensing, Wuhan University, Wuhan 430079, China (e-mail: meiqi.hu@whu.edu.cn; chen.wu@whu.edu.cn; zlp62@whu.edu.cn).

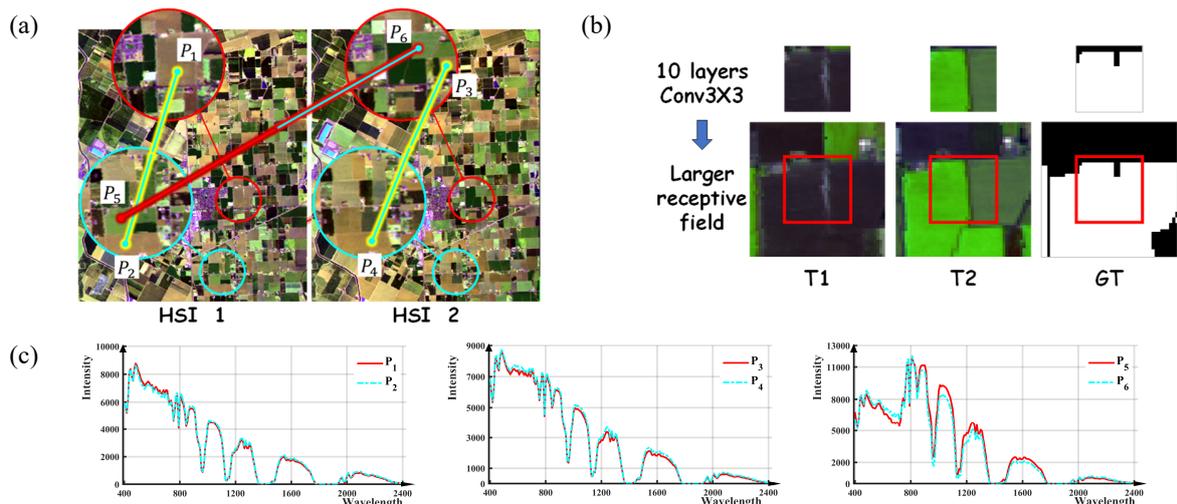

Fig. 1 (a) The illustration of the widespread local and global land cover correlations in multi-temporal hyperspectral images; (b) Local context limits the capture of integrate change boundary; (c) The corresponding to spectral curves comparison from multi-temporal HSIs of subfigure (a), where $P_1$ and $P_2$ are from HSI at time 1, $P_3$ and $P_4$ are from at HSI at time 2, $P_5$ and $P_6$ are from HSI at time 1 and time 2, respectively.

guided attention module. Additionally, Shi et.al [42] designed a multipath convolutional long short-term memory neural network (MP-ConvLSTM) which preserved time-dependent multiscale representative features while extracting spatial–spectral feature. With the carefully designed CNN and RNN networks, the methods mentioned above have led to change detection performance improvements. But the innate fixed receptive field of convolutional operation makes it intractable to fully utilize the correlations and impacts of spatial data on a larger or even global scale.

Specifically, since local spatial information primarily focuses on the details of specific regions, the network may get limited performance of understanding the variant spatial scales to some extent, lacking awareness of larger scale ground objects and corresponding changes. Moreover, the extracted features and information can be fragmented and independent among distant but correlated regions, leading to incomplete and inconsistent change detection results. Fig. 1 (a)(c) illustrates the widespread local and global land cover correlations in multi-temporal hyperspectral images. Fig. 1 (b) gives an intuitive visualization of key support of larger receptive field on identifying the complete and continuous change area.

Recently, transformer [43] has achieved significant breakthroughs in natural language processing [44]–[46] and has also been widely applied in many visual interpretation tasks of computer-vision [47]–[49] and remote sensing fields [50]–[52]. Its inherent global self-attention mechanism naturally captures long-range dependencies, which draws keen attention of researchers in remote sensing change detection. SST-Former [53] proposed a joint spectral, spatial, and temporal transformer for hyperspectral image change detection, where single spectral curves are encoded to extract spectral sequence information and the spatial transformer encoder is used to extract spatial texture information from local patches. CSA-Net [54] put forward a cross-temporal interaction symmetric attention module to integrate the difference features oriented from each temporal feature embedding. CDFormer [55] presented a transformer encoder-based HCD framework, where the bi-temporal pixel sequence within a local patch was constructed by the pixel embedding module to guide transformers to exploit change information of space and time. These methods, however, were implemented by dividing the bi-temporal HSIs into non-overlapping blocks or pixel-centered patches, where the attention was only considered in local independent self-attention block, inevitably limiting the performance of integrity of changes in the local neighborhood and sematic change connections in the global field. As a consequence, long-range dependance' is still locked in a patch level connection, which however leads to inadequate understanding of global spatial relationships and incomplete change detection map.

Inspired by this, a Global Multi-head INteractive self-attention change Detection network (GlobalMind) is proposed to dig the abundant hidden correlation between different ground objects from global view and cross-temporal view for Hyperspectral Change Detection. To obtain direct feature interactions of global image objects, we propose a simple and effective Global Axial Segmentation (GAS) strategy to directly mine long-range dependencies of the same objects and changes within the global image range. With the help of GAS, an original global spatial multi-head interactive self-attention (Global-M) module is crafted to exploit the diverse spatial correlation and discriminative representation of the same kind of objects of global scale. Meanwhile, an innovative global temporal interactive multi-head self-attention (GlobalD) module takes advantage of the natural affinity of bi-temporal images, exploring the connectivity of similar coverage transformation of local and global field. The multi-level change information acquired from shallow to deep GlobalD modules are finally fed into the classifier to yield the binary change map. The contributions of this paper can be summarized as follows:

1) To achieve precise hyperspectral change detection result, we put forward a ground-breaking Global Multi-head INteractive self-attention change Detection network (GlobalMind) which explicitly makes use of the global similarity of both single and cross-temporal hyperspectral images, successfully gaining accurate and globally coherent change detection map. The Global Axial Segmentation (GAS) strategy is significantly conductive to capture the direct dependence of global ground objects in a relatively simple way.

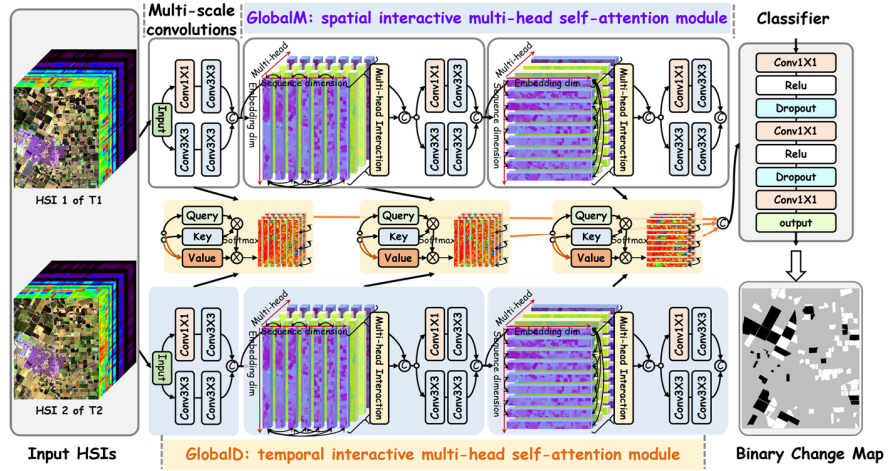

Fig. 2 The structure of proposed GlobalMind for hyperspectral change detection.

2) A novel global spatial multi-head interactive self-attention (Global-M) module is tailored to exploit the abundant potential correlations between the close neighborhood and distant outlying region from the rich and complex high-dimensional spatial-spectral space. In addition, the designed global temporal interactive multi-head self-attention (GlobalD) module combines the similarity and difference of the two temporal features to distinguish between changed and unchanged areas, and enhances the integrity and accuracy of the changed areas by strengthening the global feature interaction.

3) The superiority of proposed GloablMind on extensive experimental results and analyses on five most used real hyperspectral datasets verifies the remarkable effectiveness and efficiency on hyperspectral change detection.

The reminder of this paper is shown as follows. Section II will present detailed description of the proposed GlobalMind. And we provide the experimental results and analyses in detail on Section III. Finally, Section VI will conclude this paper.

## II. METHODOLOGY

The architecture of proposed method is shown in Fig. 2. The proposed GlobalMind firstly explores the comprehensive understanding of the global context features of each single HSI, establishing the correlation dependence of ground objects in the global scope and extracting the rich spatial and discriminant spectral features. Then the multi-level cross temporal change information is mined from the bi-temporal global spatial-spectral features, distinguishing the changed from unchanged and promoting the complete extraction of local and global similar changes. Concretely, GlobalMind is a siamese network fed with the whole multi-temporal hyperspectral images. The input is firstly processed by the multi-scale convolutions, where convolutional layers with different kernel size extract various scaled features in the local area. Then a novel GlobalM module processes the features from the integral field to explore long-range correlation, which is crucial for completeness of those similar ground objects. The extracted local and global features are fused together for feature interaction. For exploitation of diverse scaled spatial and spectral features, three creative GlobalD modules are tailored to explore the correlation of bi-temporal features from global perspective and promote the integrity of cross-temporal changes. The Global Axial Segmentation (GAS) strategy is originally designed to extend the local correlation to a large and even the global range, yielding more accurate similar changes and more complete large changes. The concatenations of multi-level temporal change features are finally fed into the Classifier to produce a precise binary change map.

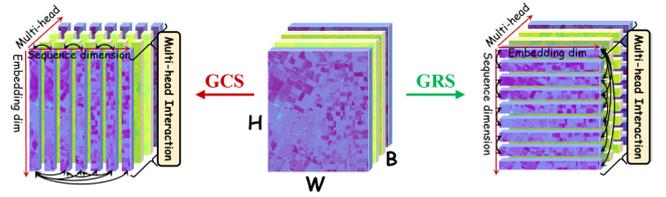

Fig. 3 The sketch map of Global Axial Segmentation (GAS) strategy, involving Global Row Segmentation (GRS) and Global Column Segmentation (GCS).

### A. Global Axial Segmentation Strategy

Self-attention mechanism naturally computes the correlation between two arbitrary tokens within a sequence, creating long-range dependance and useful for exploitation of context information. However, most of the works only consider the short-term connection of a certain local area, such as the image patch or window [53]–[55], owing to huge computation complexity of self-attention [56], [57]. Therefore, we propose the Global Axial Segmentation (GAS) (presented in Fig. 3) strategy to effectively incorporate the global context information with acceptable calculation cost.

Given an image feature map extracted from previous multi-scaled convolutions denoted as $\mathcal{F} \in \mathcal{R}^{H \times W \times B}$, where $H$, $W$, and $B$ refer to the height, width, and channel, respectively. GAS strategy directly splits the whole feature map $\mathcal{F}$ into non-overlapping sequence row by row or column by column, named as Global Row Segmentation (GRS) or Global Column Segmentation (GCS), individually. Specifically, a token is represented by a single row of feature representation among the row of spatial space for GRS. Thus, the number of tokens for GRS strategy is $H$. Meanwhile, the high-dimensional embedding representation of each row is cut into multiple sections, align with the channel dimension, which serve as various feature characteristic of a single token.

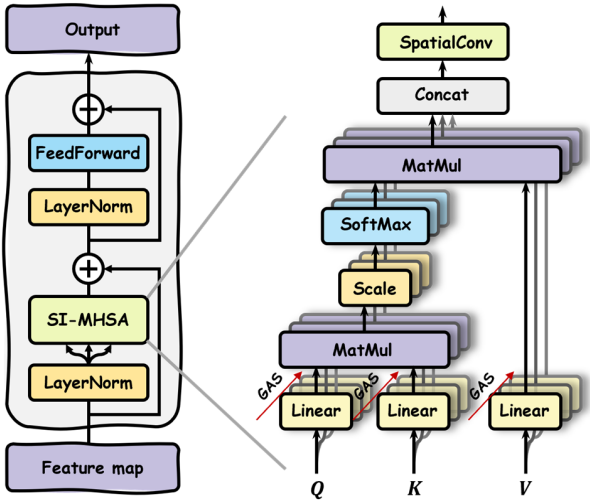

Fig. 4 The architecture of GlobalM module.

Mathematically, for GRS strategy, the input whole image feature map is transformed into a sequence of row representations $\mathcal{F} \in \mathcal{R}^{L \times N_e \times N_h}$ without rearrangement, where $L$, $N_e$, and $N_h$ are the sequence length, the num of heads, and the embedding dimension of each token, separately. The relationship can be depicted as follows:

$$L = H \quad (1)$$
$$N_e = W \quad (2)$$
$$N_h = B \quad (3)$$

The GRS strategy extends the range of sequence to the entire feature space, promoting the feature interaction throughout the global context with efficient computation capacity. In detail, the sequence length is only as short as $H$ not $H \times W$, which significantly reduces the computation complexity from $O(H^2 \times W^2)$ to $O(H^2)$ for self-attentoin. Likewise, as Fig. 3 showed, GCS strategy directly transfers the hyperspectral feature map into a column of sequence without overlapping, conveniently calculating the self-attention among the column space of entire ground object features with high efficiency. GAS strategy can be straightforwardly embedded into the GlobalM and GlobalD modules, beneficial to explore the global correlation and rich representations.

### B. GlobalM Module

Fig. 4 displays the architecture of GlobalM module. There are two sub-layers, namely the spatial interactive multi-head self-attention (SI-MHSA) and the feed forward network (FFN). The input feature map $\mathcal{F}$ is firstly processed by a LayerNorm [58] and then fed into the SI-MHSA. The linear projection, which is implemented by $1 \times 1$ convolution, encodes the feature map into three various embedding features, $Q$, $K$, and $V$, to augment diverse correlation. The GAS strategy automatically transfers the entire feature map into a limited sequence of tokens with multiple heads, easily catching the long-range objects dependence among the global context.

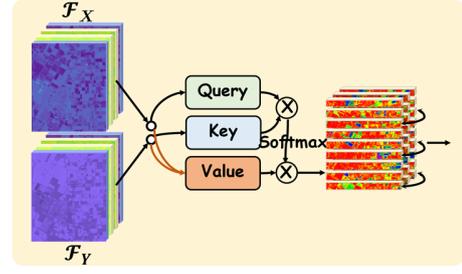

Fig. 5 The structure of proposed GlobalD module.

Attention maps between those similar ground objects located in the neighboring or distant area are generated via the scaled dot-product of the multi-head query and key features. The GAS strategy bridges the latent relevance by straightforward interactions not the information exchange from window to window. The attention map reflects the affinity of ground objects located at different regions, where the higher respondence value indicates more positive correlation. The multi-head attention maps are multiplied by the value features to obtain a weighted spatial affinity enhanced representation. The multi-head interaction is designed to improve the interaction of features of multiple heads via a light-weight $3 \times 3$ convolution, where the spatial convolution merges the diverse remote connections of different heads. The second sublayer feed forward network (FFN) is composed of two $1 \times 1$ convolution layers, used to enhance the representation ability with more powerful nonlinearity. The calculation process can be concluded as follows:

$$(Q, K, V) = (QW_Q, KW_K, VW_V) \quad (4)$$
$$\mathcal{A}_i = \textbf{softmax}\left(\frac{Q_i K_i^{\mathrm{T}}}{\sqrt{d_k}}\right), i = 1, \ldots N_h \quad (5)$$
$$\mathcal{H}_i = \mathcal{A}_i V_i \quad (6)$$
$$M_{SI-MHSA}(Q, K, V) = \omega^{3 \times 3}\left(\textbf{concat}(\mathcal{H}_1, \ldots, \mathcal{H}_{N_h})\right) \quad (7)$$

Given the input feature map $\mathcal{Z} \in \mathcal{R}^{L \times N_e \times N_h}$ fed into the GlobalM module, the mathematical computation formulas of the output $\mathcal{Z}_O$ can be depicted as:

$$\mathcal{Z}'_O = M_{SI-MHSA}(LN(\mathcal{Z})) + \mathcal{Z} \quad (8)$$
$$\mathcal{Z}_O = FFN(LN(\mathcal{Z}'_O)) + \mathcal{Z}'_O \quad (9)$$

### C. GlobalD Module

The strong correlation exists between multi-temporal features for the similar distributions of multi-temporal HSIs shot at the same area. The correlation of different positions naturally reflects the intensity of change. Inspired by this, a novel GlobalD module is proposed to explore the cross-temporal relevance and changes. And the multi-temporal spatial-spectral features of different levels are significant for dense prediction task like change detection [38], due to complex and diverse ground objects. Therefore, GlobalD module takes all the geomatic features from shallow layers and semantic features from deep layers into consideration to obtain precise and complete change detection map.

Fig. 5 shows the details of proposed GlobalD module. Given the spatial-spectral features from time 1 and time 2, denoted as $\mathcal{F}_X$ and $\mathcal{F}_Y$, separately. Differing from the standard self-attention, the GlobalD module employs $\mathcal{F}_X$ as query and the $\mathcal{F}_Y$ as key, trying to model the positional relevance which hints the change information. Moreover, the difference map of multi-temporal features directly reflects the change magnitude and are regarded as the value, which is multiplies by the attention map. The calculation formulas of query, key, and value are depicted as follows:

$$Q_t = \mathcal{F}_X W_t^Q \quad (10)$$
$$K_t = \mathcal{F}_Y W_t^K \quad (11)$$
$$V_t = \mathbf{abs}(\mathcal{F}_X - \mathcal{F}_Y) W_t^V \quad (12)$$

where $W_t^Q$, $W_t^K$, and $W_t^V$ are linear projection weight of $1 \times 1$ convolution used in temporal interactive self-attention module.

TABLE I
THE NUMBER OF TRAINING SAMPLES FOR FIVE DATASETS

| Dataset | | Hermiston | Farmland | River | Bay | Barbara |
|---|---|---|---|---|---|---|
| Sample | Unchanged | 500 | 500 | 500 | 500 | 500 |
| | Changed | 500 | 500 | 500 | 500 | 500 |
| Image Size | | 307×241 | 450×140 | 463×241 | 600×500 | 984×740 |
| Pixel for Evaluation | | 73987 | 63000 | 111583 | 73481 | 132552 |
| Percentage | | 1.3516% | 1.5873% | 0.8962% | 1.3609% | 0.7544% |

The affinity map between two features gives a weighted representation highlighting the area with strong relevance. The proposed general GAS policy is also applied to the GlobalD module perfectly, empowering the connection of similar but distant changes and accurate capture of large-scaled transformation from global view. Each head of the multiple heads can be mathematically expressed as follows:

$$\mathcal{H}_t^i = \mathbf{softmax}\left(\frac{Q_t^i(K_t^i)^\mathrm{T}}{\sqrt{d_{K_t}}}\right) V_t^i, i = 1, \dots N_h \quad (13)$$

where the final output of the temporal interactive multi-head self-attention mechanism is constructed by $3 \times 3$ convolutional fusion of multiple head representations.

$$M_{TI-MHSA}(\mathcal{F}_X, \mathcal{F}_Y) = \omega^{3 \times 3}\left(\mathbf{concat}(\mathcal{H}_t^1, \dots \mathcal{H}_t^{N_h})\right) \quad (14)$$

Given the input feature $\mathcal{F}_X$ and $\mathcal{F}_Y$ fed into the GlobalD module, the mathematical computation formulates of the output can be summarized as:

$$\mathcal{Z}_t' = M_{TI-MHSA}(LN(\mathcal{F}_X, \mathcal{F}_Y)) + \mathbf{abs}(\mathcal{F}_X - \mathcal{F}_Y) \quad (15)$$
$$\mathcal{Z}_t = FFN(LN(\mathcal{Z}_t')) + \mathcal{Z}_t' \quad (16)$$

### D. Loss Function

The GlobalD module establishes the correlation between bi-temporal spatial-spectral features by temporal-interactive multi-head self-attention. Remember that the self-attention mechanism can be regarded as a directed graph model, where the connection between two nodes should be symmetric due to the nodes of the same semantics sharing equal weight [59]. Therefore, the bi-temporal features from the two siamese branches are directly swapped when fed into the GlobalD module. And we can get two binary change detection maps $\hat{\phi}_1$ and $\hat{\phi}_2$ by such transformation, which should be symmetric through the network optimization. The cross-entropy loss is employed as the binary classification function.

$$\mathcal{L}_{bcd} = \frac{1}{N}\sum_i -(y_i \log \hat{\phi}_1) + (1 - y_i)\log(1 - \hat{\phi}_1) + \frac{1}{N}\sum_i -(y_i \log \hat{\phi}_2) + (1 - y_i)\log(1 - \hat{\phi}_2) \quad (17)$$

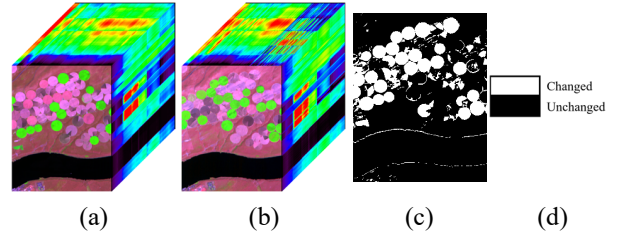

(a) (b) (c) (d)

Fig. 6 The pseudo-color map of Hermiston dataset, (a) hyperspectral image acquired at time 1, (b) hyperspectral image acquired at time 2, (c) reference map, (d) legend.

where $y$ refers to the binary change label and $N$ is the total number of training samples.

### III. EXPERIMENTS AND ANALYSIS

To testify the performance of proposed GlobalMind on hyperspectral change detection, extensive hyperspectral experiments have been conducted on five real-world datasets. In this section, we firstly provide the detailed descriptions of the datasets. Second, the details of the experimental setting are given. Third, we compare the experimental results with other state-of-the-art change detection approaches. Fourth, the discussion about the GAS strategy is presented. Fifth, the ablation test of proposed method is analyzed. Finally, we presented the running time cost to test the efficiency of GlobalMind.

#### A. Hyperspectral Datasets

Five hyperspectral change detection datasets have been opted to evaluate the effectiveness, namely, Hermiston, Farmland, River, Bay Area, and Santa Barbara datasets. And TABLE I gives a thorough statistical description of all five datasets.

*1) Hermiston dataset:* As Fig. 6 presents, the two HSIs were acquired on May 1, 2004, and May 8, 2007 by Hyperion in Hermiston city, respectively. And Fig. 6 (c) is the reference map. The scene covers a wide range of irrigated fields, river, cultivated land, with size as 307 × 241 pixels and 154 channels.

*2) Farmland dataset:* Fig. 7 shows the bi-temporal Farmland dataset shot on May 3, 2006 and April 23, 2007 by Hyperion, in the city of Yuncheng, Jiangsu, China over farmland area. The image comprises a size of 450 × 140, and 155 spectral bands.

*3) River dataset:* As shown in Fig. 8, the two HSIs were acquired at May 3, 2013 and December 31, 2013, separately, in Jiangsu province, China. This data set has a size of 463 × 241 pixels with 198 bands available after noisy band removal. The main change type is disappearance of substance in river.

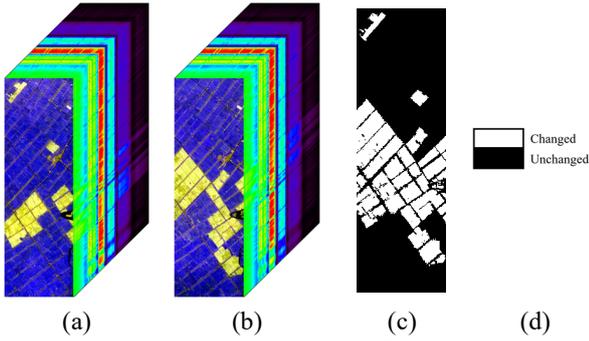

Fig. 7 The pseudo-color map of Farmland dataset, (a) hyperspectral image acquired at time 1, (b) hyperspectral image acquired at time 2, (c) reference map, (d) legend.

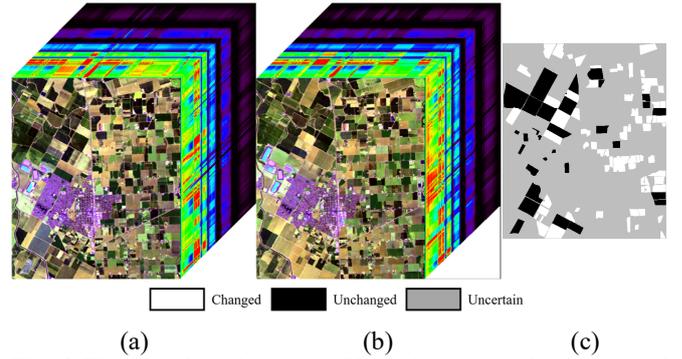

Fig. 9 The pseudo-color map of Bay dataset, (a) hyperspectral image acquired at time 1, (b) hyperspectral image acquired at time 2, (c) reference map, (d) legend.

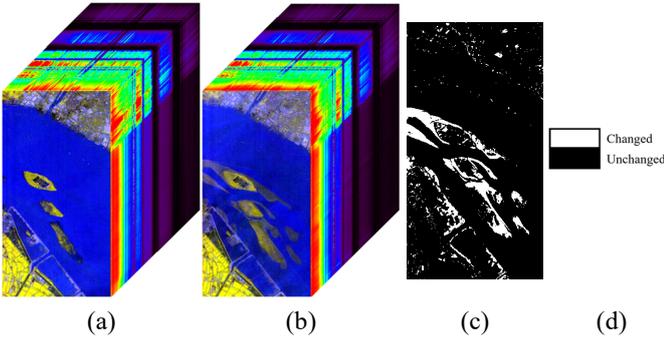

Fig. 8 The pseudo-color map of River dataset, (a) hyperspectral image acquired at time 1, (b) hyperspectral image acquired at time 2, (c) reference map, (d) legend.

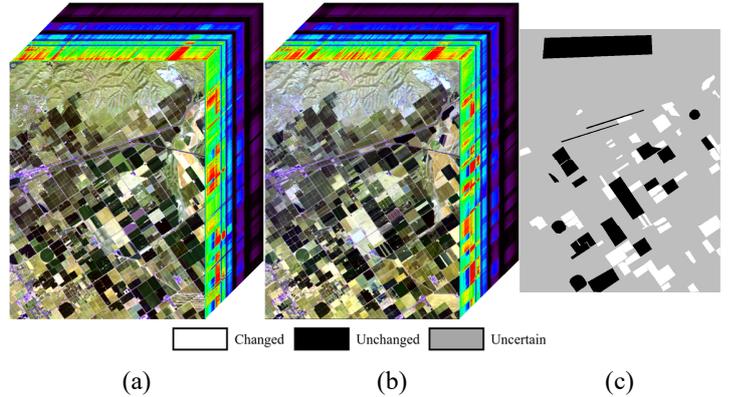

Fig. 10 The pseudo-color map of Barbara dataset, (a) hyperspectral image acquired at time 1, (b) hyperspectral image acquired at time 2, (c) reference map, (d) legend.

*4) Bay Area dataset*: Fig. 9 gives three-dimensional pseudo-color cube representation of Bay Area dataset, which are taken on 2013 and 2015, individually, with the AVIRIS sensor surrounding the city of Patterson (California). Bay dataset is largely covered by farm lands and buildings, with a large spatial size as 600 × 500 pixels and 224 spectral bands. Noted that only the labeled changed and unchanged area are adopted for assessment.

*5) Santa Barbara dataset*: The fifth one is a large-scale real dataset, exhibited as Fig. 10, where (a) and (b) are shot on the years 2013 and 2014 with the AVIRIS sensor on the Santa Barbara region. The spatial dimensions are 984 × 740 pixels and both have 224 spectral bands. The two HSIs have recorded the urban evolution and dynamic changes of farmland, and hyperspectral change detection provides a powerful tool to detect the accurate dynamics changes of urban development.

### B. Implementation Details

*1) Experimental Settings:* The experiments were implemented on a single NVIDIA RTX 3090 GPU. We adopted He-normal way [60] to initialize our method. The optimizer is Adam with L2 normalization efficient as 0.001. The initial learning rate is set as 5e-4 and a cosine decay [61] strategy is employed. We set number of total epochs for training as 200. Batch size is not required due to since the full image is fed into the proposed method. To better test the performance, a limited with SST-Former [53]. The number of training samples and the percentage have been listed in TABLE I. For GRS strategy, the Hermiston, Farmland, and River datasets employ the combination of GRS-GRS for the two GlobalM module, while number of samples have been selected from the reference, including 500 unchanged pixels and 500 changed pixels, same GCS-GRS combinations is applied to the Bay and Barbara datasets. The corresponding GlobalD module adopts the same GAS strategy with the GlobalM module to set the alignment of the extracted spatial-spectral features with the temporal change information. And the first GlobalD module models the global change information via GRS strategy. Noted that there are two binary change maps corresponding to the two sub-losses, where these two maps are supposed to be systematic according to the sematic consistency of self-attention in a directed graph [59], which is described in the loss function of part II. Therefore, the first change probability map is adopted as the final result for simplicity. In addition, the large-scale Bay dataset is segmented into upper and lower parts while the Barbara dataset is divided into four parts with each part set as half the size of height and width of original size.

*2) Evaluation Criteria:* The five most common index are used for quantitative assessment: Overall Accuracy (OA), Kappa Coefficient, F1 score, Precision and Recall rate. In more details, the detection maps are segmented into four parts for direct visualization contrast, namely, True Positive (TP), True Negative (TN), False Negative (FN) and False Positive (FP).

*3) Comparison Methods:* Eight competitive algorithms are opted for comparison and verify the effectiveness of GlobalMind, involving CVA [62], Iterative Slow Feature

TABLE II
THE DETAILS OF PARAMETER SETTINGS OF COMPARATIVE METHODS

| Parameter | GETNET | ML-EDAN | SST-Former | CSA-Net | EMS-Net | BIT | GlobalMind |
|---|---|---|---|---|---|---|---|
| Batch Size | 96 | 16 | 64 | 16 | \ | \ | \ |
| Epoch | 1000 | 200 | 400 | 100 | 200 | 100 | 200 |

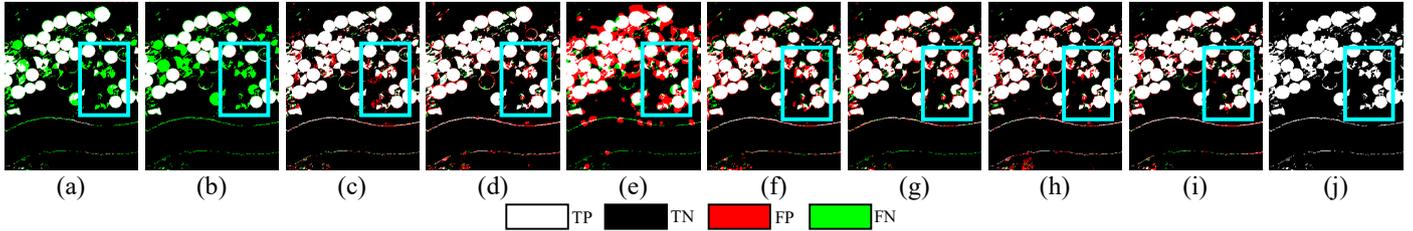

(a) (b) (c) (d) (e) (f) (g) (h) (i) (j)

TP   TN   FP   FN

Fig. 11 The binary change detection results of Hermiston dataset. (a) CVA, (b) ISFA, (c) GETNET, (d) ML-EDAN, (e) BIT, (f) CSA-Net, (g) EMS-Net, (h) SST-Former, (i) Proposed GlobalMind, (j) Reference change map.

TABLE III
THE QUANTITATIVE EVALUATION ON HERMISTON DATASET

| Method | OA | Kappa | F1 | Precision | Recall |
|---|---|---|---|---|---|
| CVA | 0.9272 | 0.7670 | 0.8103 | 0.6898 | *0.9819* |
| ISFA | 0.9023 | 0.6716 | 0.7262 | 0.5750 | **0.9852** |
| GETNET | *0.9546* | *0.8756* | *0.9054* | *0.9638* | 0.8536 |
| ML-EDAN | 0.9471 | 0.8557 | 0.8903 | 0.9529 | 0.8356 |
| BIT | 0.8392 | 0.6070 | 0.7128 | 0.8850 | 0.5969 |
| CSA-Net | 0.9285 | 0.8060 | 0.8529 | 0.9195 | 0.7954 |
| EMS-Net | 0.9293 | 0.8105 | 0.8567 | 0.9204 | 0.8066 |
| SST-Former | 0.9449 | 0.8518 | 0.8880 | **0.9697** | 0.8190 |
| GlobalMind | **0.9556** | **0.8781** | **0.9071** | 0.9620 | 0.8582 |

FOR CONVENIENCE: **BEST** AND *2nd-best*.

Analysis (ISFA) [63], GETNET [34], ML-EDAN [41], Bitemporal Image Transformer (BIT) [64], CSA-Net [54], EMS-Net [65], and SST-Former [53], covering most of the state-of-the-art CNN–, RNN-, transformer-based, and hybrid architecture-based approaches.

GETNET (without unmixing) [34] is an end-to-end convolutional neural network, incorporating a novel difference affinity of bi-temporal HSIs to provide more abundant cross-channel gradient information. The epoch for training is set as 1000, same with the setting of SST-Former and other experimental parameters are consist with the original paper'.

ML-EDAN [41] firstly presents a convolutional encoder–decoder backbone to exploit and fuse the hierarchical features and then analyze multilevel temporal dependence via the designed long short-term memory (LSTM) subnetwork.

BIT [64] argues that high-level concepts of the change of interest can be represented by a few visual sematic tokens and proposes a bi-temporal transformer encoder-decoder framework to model the long-range spatial context and obtain the change information from differenced deep features.

CSA-Net [54] put forward e an original cross-temporal interaction symmetric attention network to dig the difference features oriented from each temporal feature embedding.

EMS-Net [65] presents an Efficient Multi-temporal Self-attention Network to cut redundancy of those similar and containing-no-changes feature maps, computing efficient multi-temporal change information for precise binary result.

SST-Former [53] designs a joint spectral, spatial, and temporal transformer, where the spectral-spatial transformer is tailored to extract the spectral sequence information and spatial texture feature from patch pairs and temporal transformer to detect the change information via cross attention mechanism.

Noted that CSA-Net and SST-Former receive patch pairs for network training. All deep learning-based methods share the same number of training samples and repeated for ten times. The detailed parameter setting of these methods are summarized in TABLE II.

### C. Change Detection Results and Analysis

The change detection maps of Hermiston dataset with regard to the comparative SOTA methods and proposed GlobalMind are presented in Fig. 11. For convenience of visualization contrast, we set TP is in white, TN is in black, FP IS in Red, and FN is in green, separately. Obviously, larger number of changed area are misclassified as unchanged, which are stressed in green, in the change results of CVA and ISFA. In the result of BIT (Fig. 11 (e)), the big growing change area leads to speckles and false detections, which can be explained by the spatial information loss when the limited tokens are then resampled to the original image space under a small number of samples. As the bright blue box showed, most of the other methods can detect the circular and tiny irregular irrigated land transformations. Nonetheless, the noise and red false detection around the dense change areas can be found in the maps of comparative methods. By contrast, the proposed GlobalMind obtains more distinct and accurate change boundaries, with less green missions and red false alarms, demonstrating the validity of global sematic correlation modeling of GlobalMind on single and cross temporal HSIs. The quantitative assessment (showed in TABLE III) also proves the stronger performance of proposed method among the SOTA approaches, where the GlobalMind achieves the top OA, Kappa, and F1 score, which is highlighted in red bold.

Fig. 12 shows the change detection results on Farmland dataset. The composition of the Farmland dataset is simple, but there are many border areas where farmland and ridges intersect, which also brings challenges to change detection. Specifically, CVA, ISFA, and GETNET do not work well at the edge of

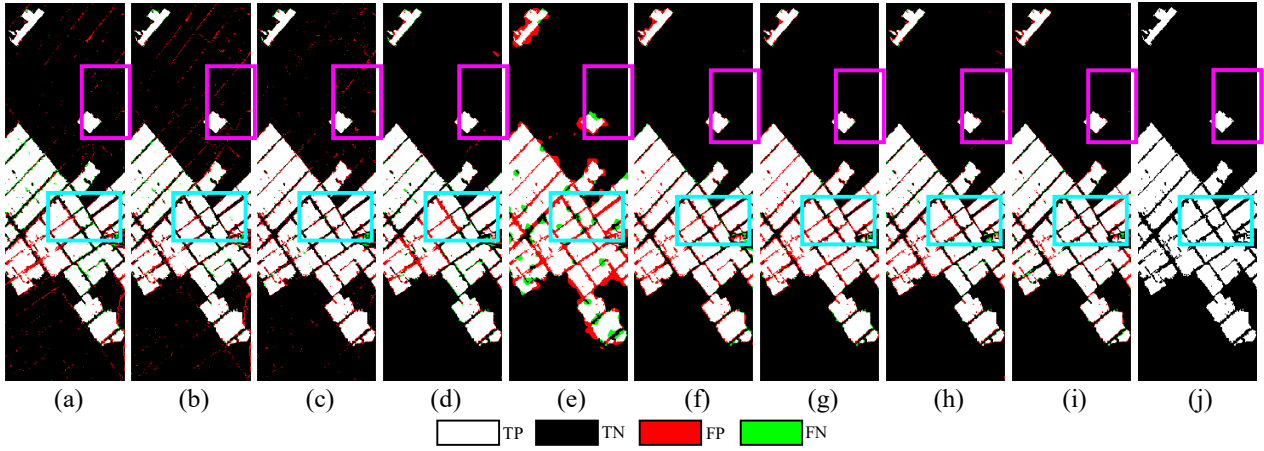

Fig. 12 The binary change detection results of Farmland dataset. (a) CVA, (b) ISFA, (c) GETNET, (d) ML-EDAN, (e) BIT, (f) CSA-Net, (g) EMS-Net, (h) SST-Former, (i) Proposed GlobalMind, (j) Reference change map.

TABLE IV
THE QUANTITATIVE EVALUATION ON FARMLAND DATASET

| Method | OA | Kappa | F1 | Precision | Recall |
|---|---|---|---|---|---|
| CVA | 0.9548 | 0.8927 | 0.9249 | 0.9600 | 0.8923 |
| ISFA | 0.9575 | 0.8996 | 0.9301 | 0.9752 | 0.8889 |
| GETNET | 0.9658 | *0.9193* | *0.9439* | 0.9015 | **0.9906** |
| ML-EDAN | 0.9635 | 0.9132 | 0.9393 | 0.9726 | 0.9081 |
| BIT | 0.9021 | 0.7746 | 0.8456 | 0.7796 | 0.9238 |
| CSA-Net | 0.9592 | 0.9034 | 0.9327 | 0.9742 | 0.8947 |
| EMS-Net | 0.9568 | 0.8986 | 0.9297 | 0.8807 | *0.9845* |
| SST-Former | *0.9659* | 0.9191 | 0.9434 | *0.9792* | 0.9102 |
| GlobalMind | **0.9700** | **0.9284** | **0.9498** | **0.9795** | 0.9219 |

FOR CONVENIENCE: **BEST** AND *2nd-best*.

TABLE V
THE QUANTITATIVE EVALUATION ON RIVER DATASET

| Method | OA | Kappa | F1 | Precision | Recall |
|---|---|---|---|---|---|
| CVA | 0.9267 | 0.6575 | 0.6955 | **0.9627** | 0.5444 |
| ISFA | 0.9341 | 0.6710 | 0.7061 | 0.9107 | 0.5765 |
| GETNET | 0.9155 | 0.6222 | 0.6652 | 0.5092 | 0.9619 |
| ML-EDAN | 0.9366 | 0.6833 | *0.9643* | 0.9378 | *0.9923* |
| BIT | 0.7995 | 0.3443 | 0.4300 | 0.2860 | 0.8681 |
| CSA-Net | 0.8992 | 0.5604 | 0.6112 | 0.9080 | 0.4612 |
| EMS-Net | 0.7761 | 0.3467 | 0.4351 | 0.2822 | 0.9702 |
| SST-Former | *0.9441* | *0.7166* | **0.9686** | *0.9439* | **0.9946** |
| GlobalMind | **0.9605** | **0.7690** | 0.7906 | 0.8574 | 0.7335 |

FOR CONVENIENCE: **BEST** AND *2nd-best*.

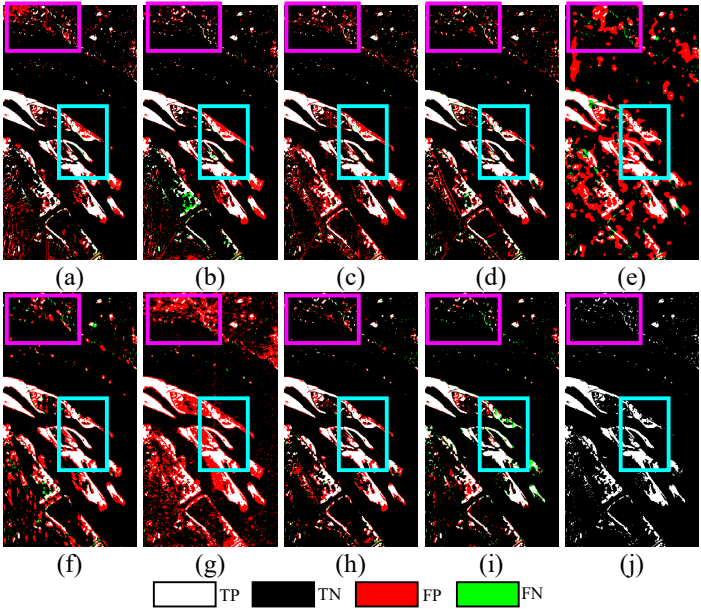

Fig. 13 The binary change detection results of River dataset. (a) CVA, (b) ISFA, (c) GETNET, (d) ML-EDAN, (e) BIT, (f) CSA-Net, (g) EMS-Net, (h) SST-Former, (i) Proposed GlobalMind, (j) Reference change map.

narrow ridge due to limited exploitation of spatial information. Some spotty red false detections can be found in the maps of the ML-EDAN, BIT, and SST-Former algorithm. On the contrary, GlobalMind can get the most desirable detection map where few pixels are mis-classified even in the center-intensive change area. For quantitative evaluation, OA, Kappa, F1, Precision, and Recall are reported in the TABLE IV. The largest value is highlighted in red bold and the second-largest is stressed in italic bold. GlobalMind obtains the best OA as 0.9700, and Kappa as 0.9284, followed by the SST-Former, which gets the second-best OA as 0.9659, and GETNET acquires the second-highest Kappa as 0.9193, confirming the superiority of proposed method.

The challenging River dataset is composed of dense urban areas and a broad river, where there are large-scale river channel changes and numerous small fine-grained changes. As Fig. 13 depict, massive red false detections can be found in the maps of most of comparative algorithms, especially for the BIT, CSA-Net, and EMS-Net. In addition, large number of noises appear in the top and bottom dense urban area for CVA, GETNET, and ML-EDAN. There are also many swelling small-scale changes in the result of CSA-Net, which origin from the emergence of boats. On the contrary, little red mis-classifications and green missed detections can be noticed in the change detection map of proposed method. The proposed GlobalM and GlobalD module helps find similar features and discover correlations between features from global perspective, which is vital for detecting integral changes and separating changes from unchanges, particularly in those complex scenes. In summary, GlobalMind obtains more precise and smooth change detection result than the other comparative methos. TABLE V lists the quantitative results. GlobalMind reports remarkable OA, Kappa than other methods do, gaining a 5% increasement of Kappa than the second-best one.

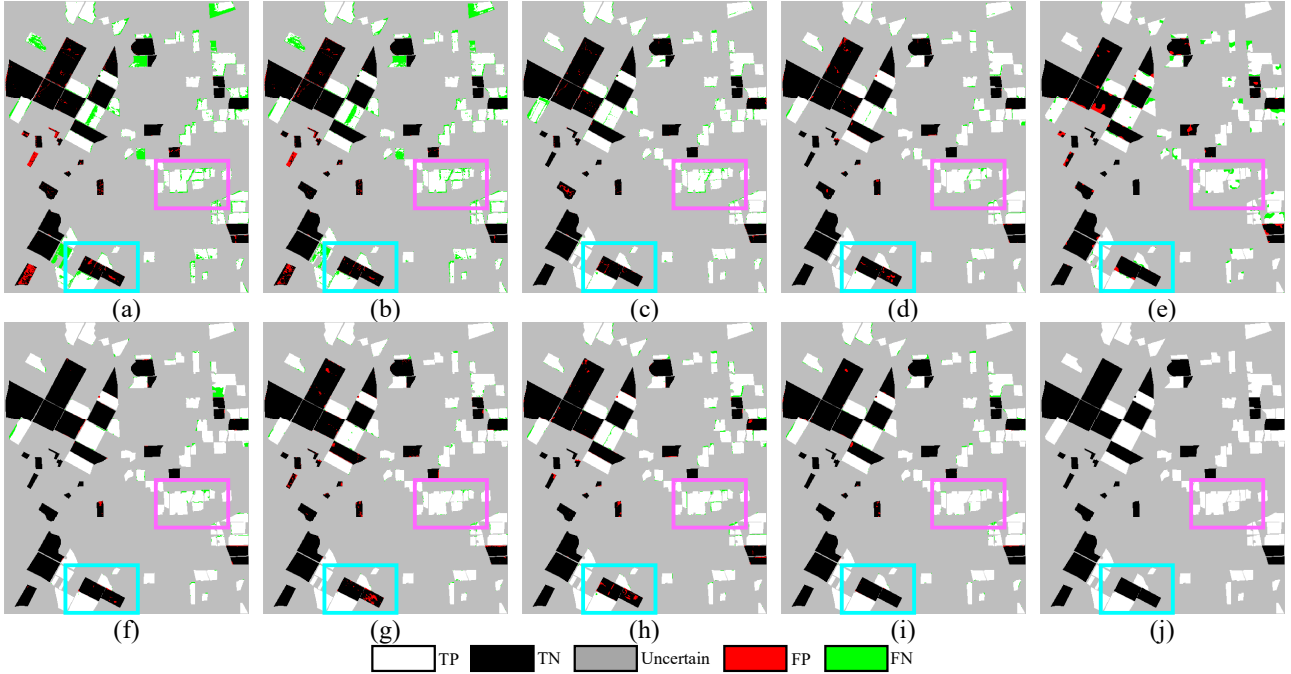

Fig. 14 The binary change detection results of Bay Area dataset. (a) CVA, (b) ISFA, (c) GETNET, (d) ML-EDAN, (e) BIT, (f) CSA-Net, (g) EMS-Net, (h) SST-Former, (i) Proposed GlobalMind, (j) Reference change map.

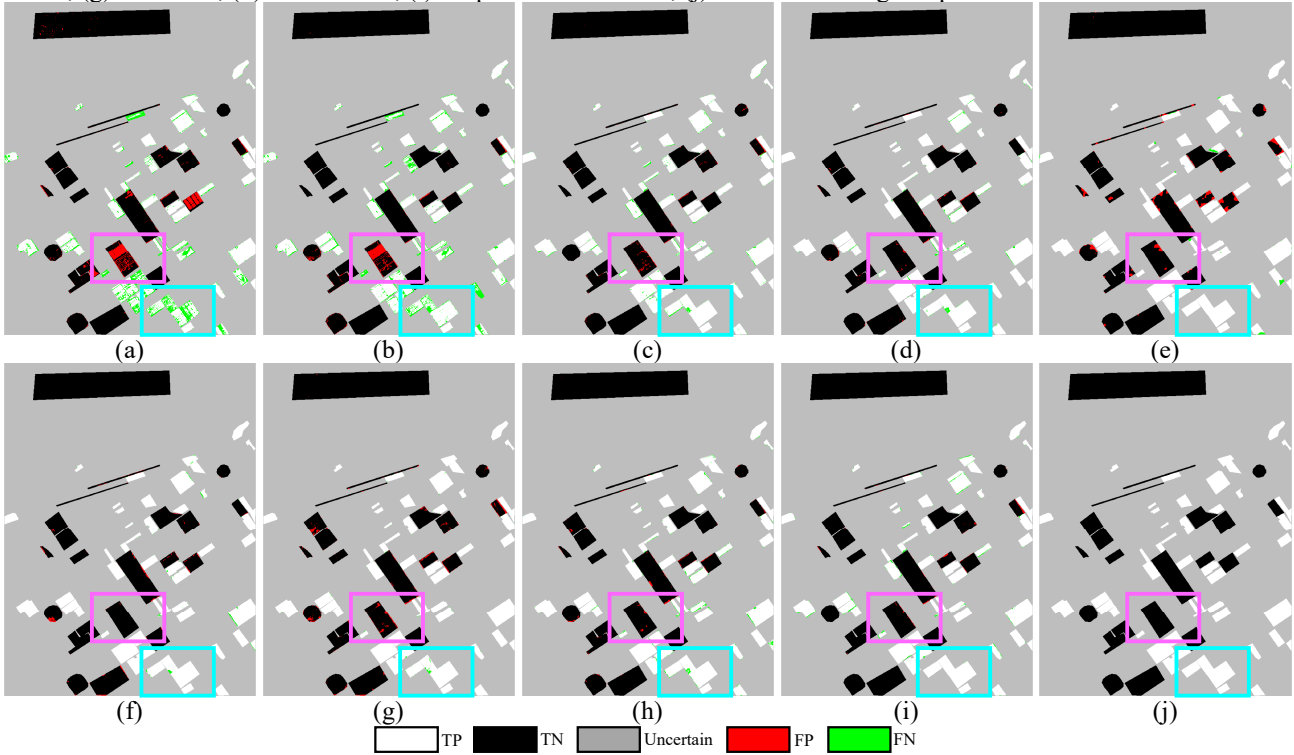

Fig. 15 The binary change detection results of Santa Barbara dataset. (a) CVA, (b) ISFA, (c) GETNET, (d) ML-EDAN, (e) BIT, (f) CSA-Net, (g) EMS-Net, (h) SST-Former, (i) Proposed GlobalMind, (j) Reference change map.

Fig. 14 and Fig. 15 represent the change detection maps of another two large-scale datasets. And the corresponding quantitative evaluation are summarized in TABLE VI and TABLE VII. The areas in grey are not labeled, marked as the uncertain. It can be seen that the maps of CVA and ISFA suffer

TABLE VI
THE QUANTITATIVE EVALUATION ON BAY DATASET

| Method | OA | Kappa | F1 | Precision | Recall |
|---|---|---|---|---|---|
| CVA | 0.8723 | 0.7462 | 0.8709 | 0.8057 | 0.9474 |
| ISFA | 0.8917 | 0.7848 | 0.8905 | 0.8234 | 0.9695 |
| GETNET | 0.9587 | 0.9173 | 0.9607 | 0.9435 | 0.9786 |
| ML-EDAN | 0.9724 | 0.9447 | 0.9741 | 0.9685 | 0.9797 |
| BIT | 0.9437 | 0.8869 | 0.9469 | 0.9396 | 0.9545 |
| CSA-Net | *0.9768* | *0.9534* | *0.9780* | *0.9688* | *0.9875* |
| EMS-Net | 0.9754 | 0.9506 | 0.9767 | 0.9686 | 0.9852 |
| SST-Former | 0.9706 | 0.9410 | 0.9723 | 0.9660 | 0.9787 |
| GlobalMind | **0.9815** | **0.9629** | **0.9826** | **0.9758** | **0.9894** |

FOR CONVENIENCE: **BEST** AND *2nd-best*.

TABLE VII
THE QUANTITATIVE EVALUATION ON BARBARA DATASET

| Method | OA | Kappa | F1 | Precision | Recall |
|---|---|---|---|---|---|
| CVA | 0.8780 | 0.7403 | 0.8376 | 0.7997 | 0.8792 |
| ISFA | 0.8912 | 0.7675 | 0.8535 | 0.8059 | 0.9071 |
| GETNET | 0.9735 | 0.9443 | 0.9659 | 0.9556 | 0.9765 |
| ML-EDAN | 0.9828 | 0.9639 | 0.978 | 0.9735 | *0.9826* |
| BIT | 0.9693 | 0.9361 | 0.9619 | *0.9841* | 0.9408 |
| CSA-Net | 0.9861 | *0.9709* | *0.9824* | **0.9886** | 0.9764 |
| EMS-Net | *0.9788* | 0.9556 | 0.9732 | 0.9803 | 0.9668 |
| SST-Former | 0.9810 | 0.9602 | 0.9758 | 0.9718 | 0.9798 |
| GlobalMind | **0.9865** | **0.9717** | **0.9828** | 0.9806 | **0.9851** |

FOR CONVENIENCE: **BEST** AND *2nd-best*.

from severe missed detections for both Bay and Barbara datasets. Additionally, a little bit false classified changes can be captured on the results of BIT, EMS-Net, and SST-Former for Bay dataset. As a comparison, GlobalMind detects almost the changed and unchanged objects with finer and more complete changes. According to TABLE VI, the proposed method outperforms the other SOTA algorithms, gaining the best OA, Kappa, F1, Precision, and Recall for Bay dataset, validating the effectiveness of GlobalMind. For Barbara dataset, the same red mis-classifications are noticed in the maps of BIT, CSA-Net, EMS-Net, especially at the boundaries of changes. Compared with these methods, GlobalMind can acquire the most accurate change detection map, which corresponds to the highest OA, Kappa, F1, and Recall among the other eight approaches (showed in TABLE VII). To conclude, GlobalMind explores the close relevance of similar land cover types of single and bi-temporal HSIs, ranging from local neighborhood to the global distant and disconnected regions, producing intelligent change detection result among the discrete similar changes and the large-span changes.

*D. Disscussions*

To explore the effect of combinations of GAS strategy on the change detection performance, experiments have been implemented on all different scaled hyperspectral datasets. There are two GlobalM modules, yielding four combinations of the GAS, namely GRS-GRS, GRS-GCS, GCS-GRS, and GCS-GCS. Fig. 16 displays the quantitative comparison results on five datasets. And Fig. 16 (f) shows the pseudo-color visualization maps of corresponding HSIs at time1. According to Fig. 16 (a), the orange striped bar exceeds the other bars on OA, Kappa, F1, and Recall, indicating the greater effect of combination of GRS-GRS on the others. The similar situation can also be found on Farmland and River datasets Fig. 16 (b)(c), as illustrated by the red boxes. However, the cases are different for Fig. 16 (d)(e). It is observed that the GCS-GRS strategy depicted in yellow bar is higher than the others for both Bay and Barbara datasets. The results show that for narrow datasets, where the number of rows is much larger than the number of columns, like Farmland, River, and Hermiston datasets, GRS-GRS combination is preferable since the spatial resolution of global information is greatly higher when the dataset is segmented along the row than along the column. While for those datasets with almost equal height and width, such as Bay and Barbara datasets, the combination of GCS and GRS obtains superior result. Interestingly, for Bay and Barbara datasets, the kappa coefficient obtained by the combination of GCS-GRS is a little bit larger than the one from the combination of GRS-GCS. Concretely, the height is a bit larger than the height for these two datasets. Thus, GRS strategy produces longer token sequence than GCS strategy does. The last GlobalM module and GlobalD module adopting the GRS strategy enjoy more abundant sematic feature interaction, probably contributing to a slightly better result.

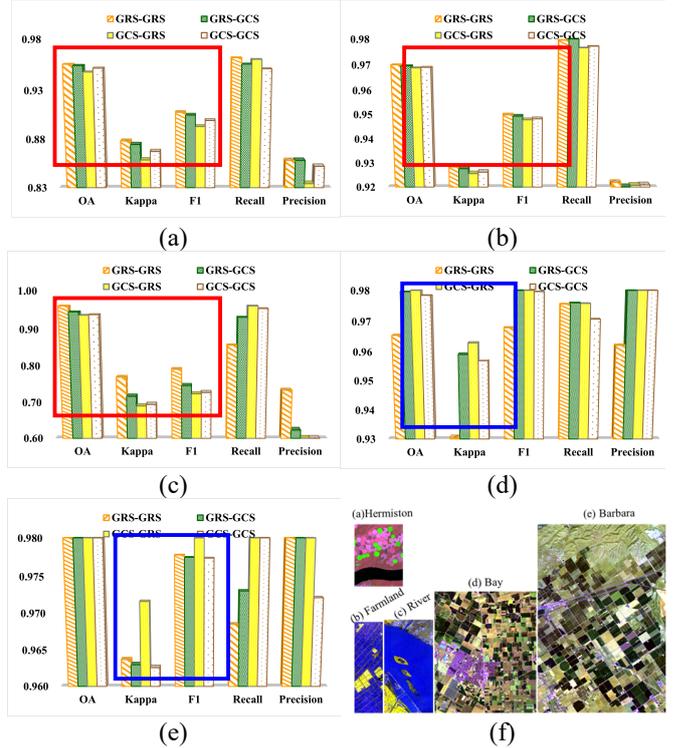

Fig. 16 The quantitative discussion results on GAS strategy on five datasets. (a) Hermiston, (b) Farmland, (c) River, (d) Bay, (e) Barbara. (f) the pseudo-color visualization of five datasets.

*E. The Ablation Study*

To evaluate the performance of each component of proposed GlobalMind, ablation studies have been conducted on five datasets. The quantitative results are listed on TABLE VIII.

TABLE VIII
THE ABLATION ASSESSMENT OF PROPOSED GLOBALMIND ON FIVE DATASETS

| Dataset | Method | GlobalM | GlobalD | OA | Kappa | F1 | Precision | Recall |
|---|---|---|---|---|---|---|---|---|
| **Hermiston** | Base | ✗ | ✗ | <u>0.8789</u> | <u>0.7023</u> | <u>0.7820</u> | **0.9634** | <u>0.6581</u> |
| | +GlobalM | ✓ | ✗ | 0.9221 | 0.7887 | 0.8398 | <u>0.9063</u> | 0.7824 |
| | +GlobalD | ✗ | ✓ | 0.9505 | 0.8635 | 0.8959 | 0.9451 | 0.8515 |
| | GlobalMind | ✓ | ✓ | **0.9556** | **0.8781** | **0.9071** | 0.9620 | **0.8582** |
| **Farmland** | Base | ✗ | ✗ | <u>0.9474</u> | <u>0.8778</u> | <u>0.9157</u> | **0.9842** | <u>0.8561</u> |
| | +GlobalM | ✓ | ✗ | 0.9512 | 0.8851 | 0.9200 | <u>0.9679</u> | 0.8767 |
| | +GlobalD | ✗ | ✓ | 0.9663 | 0.9197 | 0.9437 | 0.9728 | 0.9163 |
| | GlobalMind | ✓ | ✓ | **0.9700** | **0.9284** | **0.9498** | 0.9795 | **0.9219** |
| **River** | Base | ✗ | ✗ | <u>0.8533</u> | <u>0.4556</u> | <u>0.5235</u> | 0.9274 | <u>0.3647</u> |
| | +GlobalM | ✓ | ✗ | 0.9168 | 0.6111 | 0.6543 | 0.9059 | 0.5121 |
| | +GlobalD | ✗ | ✓ | 0.9257 | 0.6507 | 0.6894 | **0.9480** | 0.5416 |
| | GlobalMind | ✓ | ✓ | **0.9605** | **0.7690** | **0.7906** | <u>0.8574</u> | **0.7335** |
| **Bay** | Base | ✗ | ✗ | <u>0.9503</u> | <u>0.8998</u> | <u>0.9549</u> | **0.9832** | <u>0.9281</u> |
| | +GlobalM | ✓ | ✗ | 0.9706 | 0.9409 | 0.9722 | <u>0.9625</u> | 0.9821 |
| | +GlobalD | ✗ | ✓ | 0.9740 | 0.9477 | 0.9755 | 0.9699 | 0.9811 |
| | GlobalMind | ✓ | ✓ | **0.9815** | **0.9629** | **0.9826** | 0.9758 | **0.9894** |
| **Barbara** | Base | ✗ | ✗ | <u>0.9556</u> | <u>0.9083</u> | <u>0.9458</u> | **0.9840** | <u>0.9104</u> |
| | +GlobalM | ✓ | ✗ | 0.9833 | 0.9648 | 0.9785 | <u>0.9704</u> | 0.9868 |
| | +GlobalD | ✗ | ✓ | 0.9847 | 0.9679 | 0.9804 | 0.9738 | **0.9871** |
| | GlobalMind | ✓ | ✓ | **0.9865** | **0.9717** | **0.9828** | 0.9806 | 0.9851 |

FOR CONVENIENCE: **HIGNEST** AND <u>LOWEST</u>.

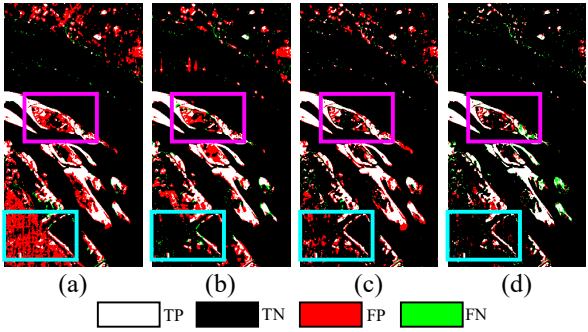

(a) (b) (c) (d)

☐ TP ■ TN ■ FP ■ FN

Fig. 17 The visualization maps of ablation comparison test on River dataset. (a) Base model; (b) Base + GlobalM module; (c) Base + GlobalD module; (d) the proposed GlobalMind.

The Base model refers to the basic framework without GlobalD and GlobalM modules. The largest values are highlighted in red bold and the smallest ones are in bold and underlined for better understanding. It is observed that both the GlobalM module and GlobalD module can significantly boost the change detection accuracy in all five datasets. Concretely, integration of GlobalD module enables model to gain more obvious improvement than GlobalD module, especially for Hermiston, Farmland, and River Datasets. GlobalD module calculates the temporal correlation maps from multi-scaled bi-temporal features, which is fused with the difference information, promoting closed relationship between the correlation and variation. With the

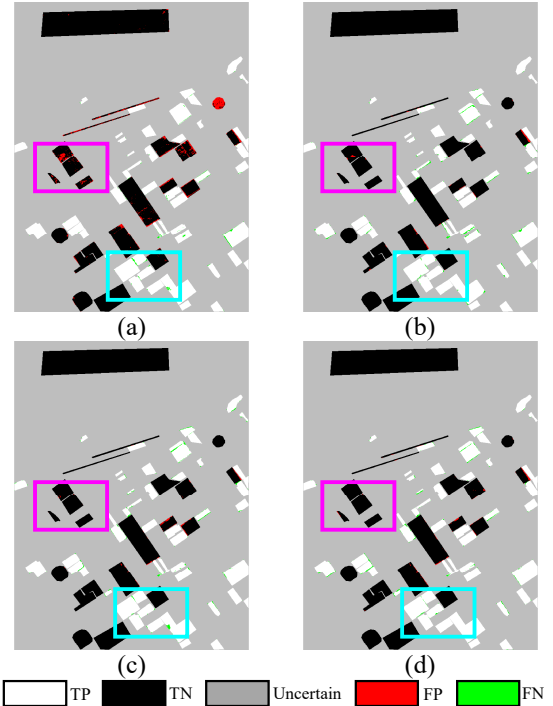

(a) (b)

(c) (d)

☐ TP ■ TN ■ Uncertain ■ FP ■ FN

Fig. 18 The visualization maps of ablation comparison test on Barbara dataset. (a) Base model; (b) Base + GlobalM module; (c) Base + GlobalD module; (d) the proposed GlobalMind.

TABLE IX
THE RUNNING TIME COST COMPARISON OF ALL METHODS ON FIVE DATASETS

| DataSet | Method | GETNET | ML-EDAN | BIT | CSA-Net | EMS-Net | SST-Former | GlobalMind |
|---|---|---|---|---|---|---|---|---|
| **Hermiston** | Total Time(s) | 854.3069 | **1171.491** | 24.3623 | 427.2919 | <u>24.3614</u> | 610.6557 | 69.6259 |
| | Single Epoch(s) | 0.7901 | 4.2591 | 0.1808 | 2.8820 | <u>0.1463</u> | **6.0996** | 0.3308 |
| | Test time(s) | **88.2466** | 84.3110 | 0.0508 | 76.8984 | <u>0.0086</u> | 49.1753 | 0.0124 |
| **Farmland** | Total Time(s) | 903.0582 | **1117.0007** | 24.1584 | 417.6745 | <u>20.2780</u> | 909.0686 | 61.8202 |
| | Single Epoch(s) | 0.8088 | **5.6160** | <u>0.1795</u> | 3.0454 | 0.2623 | 3.5988 | 0.3002 |
| | Test time(s) | 76.6026 | 72.1005 | 0.0522 | 70.8786 | <u>0.0086</u> | **80.4809** | 0.0153 |
| **River** | Total Time(s) | **1371.8155** | 1328.1943 | <u>25.2424</u> | 582.5029 | 37.0047 | 638.7135 | 105.8417 |
| | Single Epoch(s) | 1.2069 | 7.6048 | <u>0.1749</u> | 3.2008 | 0.2022 | **8.8260** | 0.5144 |
| | Test time(s) | **159.1276** | 91.2390 | 0.0562 | 133.7027 | <u>0.0087</u> | 68.0156 | 0.0122 |
| **Bay** | Total Time(s) | **2324.9187** | 1250.8948 | <u>35.5644</u> | 756.6182 | 99.1925 | 955.2522 | 210.0112 |
| | Single Epoch(s) | 1.5362 | 4.6278 | <u>0.1825</u> | 3.4629 | 0.6200 | **6.3957** | 0.9674 |
| | Test time(s) | **799.5605** | 83.6846 | 0.0643 | 92.5724 | <u>0.0096</u> | 43.4732 | 1.7594 |
| **Barbara** | Total Time(s) | **2955.7777** | 1388.9077 | <u>69.7972</u> | 1282.6321 | 237.6671 | 1122.1135 | 507.3258 |
| | Single Epoch(s) | 1.5023 | 5.2180 | <u>0.3287</u> | 3.1615 | 1.1846 | **10.0017** | 2.3480 |
| | Test time(s) | **1445.1741** | 106.5515 | 0.0792 | 150.7430 | <u>0.0101</u> | 87.9655 | 3.6890 |

FOR CONVENIENCE: **HIGNEST** AND <u>**LOWEST**</u>.

help of the GAS strategy, GlobalD module is capable of achieving change information interaction at global scope, which is critical to mine the similar change transformation of entire images and integral large-scale changes. Furthermore, the combination of GlobalM and GlobalD modules contributes to superior change detection result, where GlobalM module plays an additional role in understanding specific distribution and potential sematic consistency of ground objects from global point.

Taking the representative River and Barbara datasets as an example, the intuitive change detection maps of the ablation comparison are presented in Fig. 17 and Fig. 18, respectively. Compared with the Base model, the incorporation of the GlobalM module or GlobalD module helps model sharply reduce the red false alarms for both two datasets. Concretely, addition of GlobalM module contributes to deep interactive and discriminative spatial and spectral features, largely alleviating the variant and unsmooth spatial feature of the bottom and top urban area for River dataset (as illustrated in the blue box in the Fig. 17 (b)). Moreover, GlobalD module successfully enhances the separability between change and unchanged area, with the assistance of global change feature interaction, as the pink box exhibit in Fig. 17 (c) and Fig. 18 (c). After adding both GlobalM and GlobalD modules, GlobalMind achieves remarkably accurate change detection map for both two datasets (shown in Fig. 17 (d) and Fig. 18 (d)), making full use of the global spatial correlation of single and cross-temporal HSIs for global view.

### F. Running Time Cost Analysis

Running time test have been conducted to assess the efficiency of proposed method. TABLE IX lists the total time cost, test time, and time assumption of average single epoch for six deep learning-based methods and proposed GlobalMind.

Noted that the time of test and single epoch are of more valid reference value than the total time since different methods adopt different number of epochs. The largest value in stressed in red bold and smallest one is underlined and in bold. Specifically, it is observed that EMS-Net and BIT shares similar time cost, acquiring the lest time of single epoch and testing for five datasets. Additionally, SST-Former and GETNET require the most time for training singe epoch and testing, slowing down the efficiency. By contrast, GlobalMind demands little time for testing and share almost similar time with EMS-Net and BIT for training each epoch. To conclude, the proposed GlobalMind outperforms most of comparative methods on running time cost, with receptive time cost under outstanding change detection performance.

## V. CONCLUSIONS

In this study, an innovative Global Multitemporal INteractive self-attention Network (GlobalMind) is proposed for hyperspectral change detection, providing global sematic awareness of similar distant changes and complete large-scale changes. Specifically, GAS strategy is put forward to transform the entire hyperspectral feature cubic into a row or column sequence with limited length while remaining the spatial objects details. The designed GlobalM module extracts fine-grained spatial correlation and abundant spatial-spectral features with the combination of GAS strategy. And for better exploitation of cross-temporal relevance, GlobalD module explores the global dependance among the sematic-close object changes, promoting the accurate detection of similar changes in the global space and integral changes of a local neighborhood. On five small-scaled and large-scaled hyperspectral datasets, our method achieves superior performance among the state-of-the-art CNN-, RNN-, transformer-, and hybrid algorithms with very high efficiency.